\documentclass[conference]{IEEEtran}
\IEEEoverridecommandlockouts
\usepackage{cite}
\usepackage{amsmath,amssymb,amsfonts}
\usepackage{graphicx}
\usepackage{textcomp}
\usepackage{stfloats}
\usepackage{xcolor}
\usepackage{booktabs}
\usepackage{array}
\usepackage{float}
\usepackage{multirow}
\usepackage{url}
\usepackage[hidelinks]{hyperref}
\newcommand\blfootnote[1]{%
  \begingroup
  \renewcommand\thefootnote{}\footnote{#1}%
  \addtocounter{footnote}{-1}%
  \endgroup
}
\begin{document}

\title{Co-Evolving Policy and Reward for Code Generation: A Mechanistic Study of Single-Adapter Drift}

\author{
\IEEEauthorblockN{Amarsaikhan Batjargal}
\IEEEauthorblockA{\textit{Jarvis College of Computing and Digital Media} \\
\textit{DePaul University}\\
Chicago, IL, USA \\
abatjarg@depaul.edu}
\and
\IEEEauthorblockN{Noriko Tomuro}
\IEEEauthorblockA{\textit{Jarvis College of Computing and Digital Media} \\
\textit{DePaul University}\\
Chicago, IL, USA \\
tomuro@cs.depaul.edu}
}

\maketitle
\blfootnote{\copyright~2026 IEEE. Personal use of this material is permitted.
Permission from IEEE must be obtained for all other uses.}

\begin{abstract}
Reinforcement learning with verifiable rewards (RLVR) is particularly well
suited to code generation because generated programs can be objectively
evaluated through execution. Execution also provides behavioral information
that can be reused as additional supervision. The Synergistic Policy and Reward Co-Evolving Framework (SPARK), originally proposed for
reasoning tasks, exploits this idea by recycling training rollouts and their
evaluation results to train a model not only to generate solutions but also to
evaluate and improve them. We investigate whether this policy--reward
co-evolution approach transfers effectively to code generation and whether
execution-grounded reflection provides an additional source of
self-improvement. We compare SPARK-style rollout recycling with execution-only
Group Relative Policy Optimization (GRPO) using Qwen2.5-Coder-3B-Instruct. Our
results do not support either hypothesis under the configuration tested.
Although co-evolution improves training performance, these gains do not
generalize to held-out problems. Compared at the same stage of training, rollout
recycling produces roughly $34\times$ greater KL divergence from the reference
policy than execution-only GRPO. Over training, trained tokens per GRPO
sequence fall from 185 to 86 as training pass rate increases, while held-out
performance declines or fails to sustain early gains. We characterize these
patterns as single-adapter policy drift: the shared trainable parameters
increasingly depart from pretrained code-generation behavior without
corresponding gains in generalization. The results show that higher training
reward alone is not evidence of generalizable code self-improvement and
highlight the importance of held-out evaluation and drift diagnostics.
\end{abstract}

\begin{IEEEkeywords}
code generation, reinforcement learning, verifiable rewards, GRPO,
policy--reward co-evolution, self-correction, large language models, LoRA
\end{IEEEkeywords}

%-------------------------------------
\section{Introduction}
%-------------------------------------

Large language models (LLMs) have become increasingly capable of code
generation, but functional correctness remains challenging. A generated program
may be syntactically valid yet fail edge cases, implement an incorrect
algorithm, or crash at runtime. Because generated programs can be objectively
evaluated through execution, code generation provides a natural setting for
reinforcement learning with verifiable rewards (RLVR), unlike many
natural-language tasks that require subjective judgments of response quality.
Moreover, execution provides information about program behavior that can serve
as additional supervision.
 
This work investigates whether such information can be exploited through
policy--reward co-evolution. We build on the Synergistic Policy and Reward Co-Evolving Framework (SPARK)~\cite{spark}, a self-improvement
framework originally developed for reasoning tasks. In conventional RLVR,
model-generated solutions, or rollouts, are used to compute rewards and update
the policy, but are not typically reused for further training. SPARK instead
recycles these rollouts and their evaluation results to train the model to
evaluate and improve its own outputs. These capabilities co-evolve with the
policy within a single model. We investigate whether this approach transfers
effectively to code generation.
 
We test two hypotheses. First, we hypothesize that SPARK-style co-evolution can
improve code generation beyond an execution-only RLVR baseline based on Group
Relative Policy Optimization (GRPO)~\cite{deepseekmath}, including its ability
to generalize to problems not seen during training. Second, we hypothesize that
execution-grounded reflection, in which the model learns to revise failing code
using unit-test feedback, can provide an additional source of self-improvement.
 
We evaluate these hypotheses using the Mostly Basic Python Problems
(MBPP)~\cite{mbpp} and HumanEval~\cite{humaneval} datasets. We train on a
subset of MBPP and evaluate generalization both on held-out MBPP problems and
on HumanEval as a separate, fully held-out benchmark. This design allows us
to distinguish improvements on the training problems from in-distribution
generalization within MBPP and cross-benchmark generalization to HumanEval.
 
Our experiments do not support either hypothesis under the configuration
tested. Although co-evolution improves performance on the training problems,
these gains do not generalize to unseen problems. At the same time, the model
increasingly departs from its pretrained behavior, as measured by KL
divergence. The model also produces progressively shorter outputs during
training. We characterize these patterns as single-adapter policy drift,
arising when the same trainable parameters are used both to generate code and
to learn to evaluate it.
 
This paper makes three contributions. First, we adapt SPARK-style
policy--reward co-evolution to code generation by using information from
program execution as additional supervision. Second, we evaluate whether this
approach improves code-generation performance and generalization beyond
execution-only GRPO. Third, we analyze policy drift to examine the relationship
between training improvement and held-out generalization. Our findings
highlight limitations of the single-adapter co-evolution design studied here
and motivate further investigation of alternative designs for code generation.
%-------------------------------------
\section{Background}
%-------------------------------------
\subsection{RLVR and GRPO}
Code-generation LLMs are typically built through large-scale pretraining on
code corpora~\cite{codellama,deepseekcoder} followed by instruction tuning and
reinforcement learning from human feedback, which optimizes the model against a
reward model learned from human preferences~\cite{ouyang2022training}. RLVR
replaces that learned reward model with an executable or rule-based verifier.
DeepSeekMath introduced GRPO in this setting, and DeepSeek-R1 showed that RLVR
can produce strong reasoning behavior~\cite{deepseekmath,deepseekr1}. Related
work also improves models at inference time rather than training time, for
example by spending more computation on difficult problems~\cite{testtimecompute}.

GRPO samples several candidate solutions for the same problem and compares
their rewards within that group. For a task $x$, the model samples $K$
solutions $y_1, \ldots, y_K$, each receiving an execution reward $r_i$, and the
group-relative advantage is
\begin{equation}
A_i = \frac{r_i - \mu_r}{\sigma_r + \epsilon},
\label{eq:advantage}
\end{equation}
where $\mu_r$ and $\sigma_r$ are the mean and standard deviation of rewards in
the group and $\epsilon$ is a small constant for numerical stability. A solution
that scores above its group's average is reinforced, and one below it is
discouraged. Groups in which every solution receives the same reward carry no
learning signal and are skipped. Because ranking several candidate programs for
the same problem is exactly what a unit-test verifier provides, GRPO fits
execution feedback naturally; it is also the optimizer SPARK uses~\cite{spark}.

\subsection{Self-Correction and SCoRe}
In this paper, \emph{reflection} means training the model to repair its own
failing code: given a program that failed its unit tests and the resulting
error message, the model learns to produce a corrected program.
Self-Correction via Reinforcement Learning (SCoRe)~\cite{score} motivates this
part of our work. SCoRe trains a model in a multi-turn setting: the
model first produces an initial answer and then, conditioned on its own first
answer, produces a revised answer. SCoRe identifies two failure modes directly
relevant to execution-grounded code repair: \emph{distribution shift}, when
correction examples do not match the model's own current mistakes, and
\emph{behavior collapse}, when the model learns conservative or minimal edits
instead of meaningful repairs. These ideas help interpret our reflection
results: although we use execution traces and on-policy rollouts, reflection is
still trained as auxiliary supervised fine-tuning (SFT) rather than as a
dedicated multi-turn repair objective optimized with reinforcement learning.

\subsection{SPARK-Style Rollout Recycling}
For each generated program, execution provides correctness, the fraction of
tests passed, the error type, and the error message. SPARK~\cite{spark} turns
these into three auxiliary objectives for the same model:
\textbf{pointwise judgment} (predict whether a candidate is correct),
\textbf{pairwise comparison} (choose the better of two candidates), and
\textbf{reflection} (repair failing code). Related approaches that train a
policy together with its own reward signal include Cooper~\cite{cooper},
CURE~\cite{cure}, and self-rewarding language models~\cite{selfreward}.

In our implementation, pointwise judgment is not used (weight 0.00), because
its one-word targets such as ``Correct'' or ``Incorrect'' differ sharply from
the full programs the same model must generate. For pairwise comparison, each
failing rollout of a problem is paired with a randomly chosen passing rollout
of the same problem, with the order in which the two are shown randomized.
Reflection is the dominant auxiliary objective, since learning to repair code
from execution feedback is the main adaptation from reasoning tasks to code.
For reflection, the model is shown its failing program and the error message,
and is trained to output a correct program for the same problem: a passing
rollout from the same batch when one exists, and otherwise the canonical solution provided with MBPP.

\begin{table}[ht]
\centering
\caption{The three SPARK-style auxiliary objectives in our final co-evolution
configuration. Pointwise judgment is not used, to avoid short-target
interference with full code generation.}
\label{tab:aux-objectives}
\footnotesize
\begin{tabular}{p{1.6cm} c p{2.0cm} p{2.0cm}}
\toprule
Objective & Weight & Input & Supervised target \\
\midrule
Pointwise & 0.00 & one candidate solution & correctness label (not used) \\
Pairwise & 0.05 & two candidate solutions & the better candidate \\
Reflection & 1.00 & failing code + execution trace & correct rollout, else canonical solution \\
\bottomrule
\end{tabular}
\end{table}

Table~\ref{tab:aux-objectives} summarizes the three objectives and their
weights in our final co-evolution configuration. The weights are not
normalized to sum to one; they multiply each example's loss before training.
Setting reflection to 1.0 keeps it on the same scale as the GRPO loss, while
pairwise at 0.05 keeps it a small auxiliary signal.

\subsection{Policy Drift as a Central Risk}
We fine-tune with low-rank adaptation (LoRA)~\cite{lora}, which trains a small
set of added weights, an \emph{adapter}, while the pretrained model stays
frozen. A central risk in co-evolutionary training is that the auxiliary
objectives update the same adapter used for code generation. Their targets look
very different from ordinary code completions: a pairwise answer is a single
token, and a reflection target is a program written in response to a different
prompt. Training on this mixture can shift the adapter away from how the
pretrained model writes code. Such a shift can raise training reward without
improving, and possibly while harming, performance on unseen problems. We
track two signals of this drift. The first is the KL divergence between the
trained model and the frozen pretrained model, which measures how far its
output distribution has moved. The second is the trained tokens per GRPO sequence: the average length, in
generated tokens, of the solutions used in each GRPO update. Shorter outputs
are not harmful in themselves, but a sharp drop suggests the adapter is being
pulled toward the short auxiliary targets.
Following SPARK~\cite{spark}, we also add the KL divergence to the training
loss as a penalty; if KL still grows substantially, the penalty is not
preventing drift.

%-------------------------------------
\section{Methodology}
%-------------------------------------
\subsection{Model and Fine-Tuning Setup}
All experiments fine-tune Qwen2.5-Coder-3B-Instruct with QLoRA~\cite{qlora},
which loads the pretrained weights in compressed 4-bit form and keeps them
frozen, so that only the LoRA adapter is trained. Following the QLoRA
recommendation, the adapter is applied to all attention and feed-forward
projection layers, giving 29.9M trainable parameters out of 3.1B (about
0.96\%). We chose the 3B model so that repeated rounds of generation,
execution, and training fit within our available hardware.

The KL penalty is computed with a standard non-negative estimator, weighted by
a coefficient $\beta$, and not capped: the
model is discouraged from drifting away from the frozen pretrained model, but
is not prevented from doing so. This is deliberate, since a hard limit on drift
would hide the very effect we want to measure. Stronger constraints and early
stopping are left to future work.

\subsection{Training Objective}
Each training iteration of the co-evolution condition updates the same adapter
in two phases. The first is a standard GRPO update~\cite{deepseekmath}, which
increases the probability of solutions with positive advantage
in~\eqref{eq:advantage} and decreases it for those with negative advantage.
Each update is clipped so the model cannot change too much in a single step,
and the KL penalty with weight $\beta$ discourages drift from the frozen
pretrained model. The second phase is auxiliary supervised fine-tuning on the
recycled rollouts, minimizing
\begin{equation}
\mathcal{L}_{\mathrm{aux}}(\theta) =
\lambda_{\mathrm{aux}}\bigl(w_{\mathrm{pair}}\mathcal{L}_{\mathrm{pair}}(\theta)
+ w_{\mathrm{refl}}\mathcal{L}_{\mathrm{refl}}(\theta)\bigr),
\label{eq:aux-loss}
\end{equation}
where $\theta$ are the adapter parameters, $\mathcal{L}_{\mathrm{pair}}$ and
$\mathcal{L}_{\mathrm{refl}}$ are the standard next-token prediction losses on
the pairwise and reflection targets, computed on the target response only and
not on the prompt, and $w_{\mathrm{pair}}$ and
$w_{\mathrm{refl}}$ are the weights in Table~\ref{tab:aux-objectives}. The
overall scale $\lambda_{\mathrm{aux}}$ controls how strongly the auxiliary
phase moves the adapter relative to the GRPO phase, since both share one
optimizer. The execution-only baseline runs the GRPO phase alone; the
co-evolution condition runs both, so both update the same shared adapter,
which is the design choice under test. The values of $\beta$ and
$\lambda_{\mathrm{aux}}$ for each configuration are given in
Section~\ref{sec:results}.

\subsection{Datasets, Execution, and Reward}
We use the MBPP-sanitized subset~\cite{mbpp}, a hand-verified version of MBPP
curated by its original authors to remove ambiguous task descriptions, poorly
specified requirements, and broken test assertions. We keep the 411 problems
whose unit tests can be parsed automatically and randomly split them, with a
fixed seed, into 311 training and 100 held-out problems.
HumanEval~\cite{humaneval} (164 problems) serves as a second, fully held-out
benchmark.

To score a solution, we extract the Python code from the model's response and
run it with the problem's unit tests in a separate, isolated process. Each run
is limited to 10 seconds and 2~GB of memory, enough for even naive brute-force
solutions on these small problems while stopping infinite loops and runaway
memory use. During training, each unit test is run separately, and a
solution's reward is the proportion of its unit tests that pass, reduced by a
penalty if the code has a syntax error, raises a runtime error, or times out.

These rewards are the $r_i$ in~\eqref{eq:advantage}, computed for the $K$
candidate solutions in each GRPO group. We start with $K=4$ solutions per
problem. If all of them receive the same reward, the group carries no learning
signal, so we sample more, up to $K=8$, until the rewards differ or the cap is
reached.

\subsection{Experimental Design}
\begin{figure*}[t]
\centering
\includegraphics[width=\textwidth]{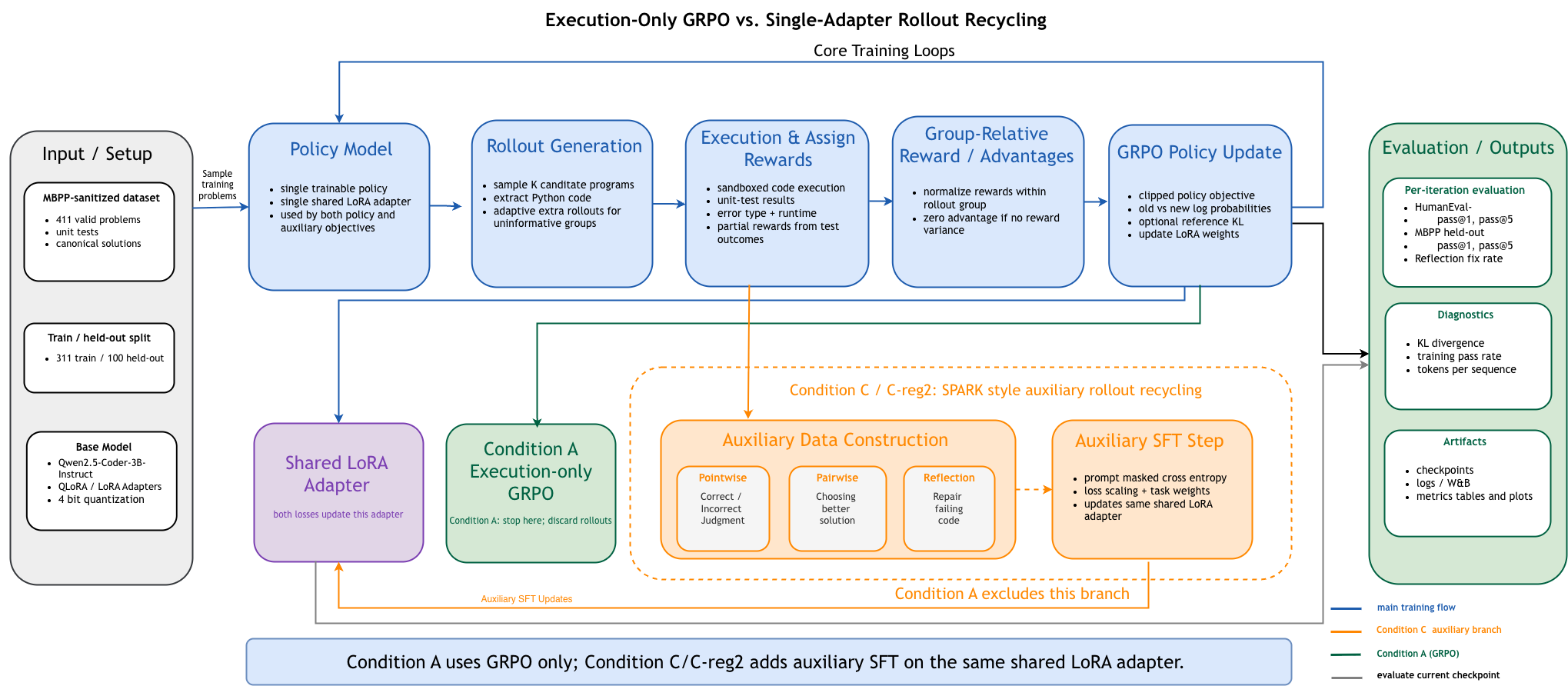}
\caption{Execution-only GRPO vs. single-adapter rollout recycling. The blue
path shows the shared execution-grounded GRPO loop used by both conditions:
MBPP problems are sampled, the policy generates candidate programs, programs
are executed in a sandbox, execution rewards are assigned, group-relative
advantages are computed, and the GRPO policy update is applied. Condition~A
stops after the GRPO update and discards rollouts. Condition~C adds the
orange auxiliary branch, where execution-labeled rollouts are recycled into
pointwise, pairwise, and reflection supervision. Both the GRPO policy update
and the auxiliary SFT step update the same shared LoRA adapter. The right-hand
boxes show per-iteration evaluation, diagnostics, and artifacts.}
\label{fig:spark-code-architecture}
\end{figure*}

We compare two conditions, depicted in
Figure~\ref{fig:spark-code-architecture}, where the blue path is shared by
both. \textbf{Condition~A} (green box) is the execution-only baseline: it runs
GRPO alone and discards its rollouts after each update. \textbf{Condition~C}
(dashed orange box) adds rollout recycling: the same rollouts are
turned into auxiliary judgment and reflection examples, and both phases update
the same shared adapter. Condition~A therefore measures what execution-only
GRPO can achieve, and Condition~C tests whether recycling improves on it. We
evaluate several settings of Condition~C that differ in how strongly the
auxiliary phase and the KL penalty are weighted; Section~\ref{sec:results}
introduces them as C-light, C-reg, and C-reg2.

\subsection{Evaluation Metrics}
We evaluate each checkpoint on two held-out sets. The 100 held-out MBPP
problems measure generalization to unseen problems from the benchmark used in
training. HumanEval, which is never used in training, measures whether training
generalizes to entirely unseen data from a separate benchmark. For each
problem, we generate five solutions at temperature 0.2 and report pass@1 and
pass@5, where pass@$k$ is the probability that at least one of $k$ solutions
passes all tests; pass@1 is thus the average fraction of the five that pass,
and pass@5 whether any of them does. HumanEval pass@1 is our primary metric,
since it measures single-attempt quality on fully unseen problems. We also
track training pass rate and mean reward (progress on the training problems),
KL divergence (policy drift), trained tokens per GRPO sequence (length drift),
and reflection fix rate (repair ability). Because all experiments use a single
seed and HumanEval has 164 problems, we do not claim statistical significance
for small pass@$k$ differences and interpret them together with the drift
diagnostics.
%-------------------------------------
\section{Results and Analysis}
\label{sec:results}
%-------------------------------------
\subsection{Preliminary Hyperparameter Study}
\label{sec:hyperparam-study}
To choose the auxiliary loss scale and KL coefficient affordably, we first ran
a shorter preliminary study: three iterations on 200 of the 311 training
problems, evaluated on the same held-out MBPP problems and on HumanEval. The study compared Condition~A, execution-only GRPO with no auxiliary phase, against two
co-evolution configurations: naive (C-light, $\beta = 0.01$,
$\lambda_{\mathrm{aux}} = 0.10$) and regularized (C-reg, $\beta = 0.02$,
$\lambda_{\mathrm{aux}} = 0.03$).
Table~\ref{tab:stage1-summary} summarizes the results.

\begin{table*}[t]
\centering
\caption{Preliminary hyperparameter study. HE = HumanEval; MBPP = held-out
MBPP; @1 and @5 denote pass@1 and pass@5; Train Pass = training pass rate;
Refl.\ Fix = reflection fix rate. Iteration~0 is the shared pretrained
baseline; metrics shown are final-iteration values. Bold marks the best
non-baseline value in each column.}
\label{tab:stage1-summary}
\footnotesize
\begin{tabular}{lcccccccc}
\toprule
Condition & HE@1 & HE@5 & MBPP@1 & MBPP@5 & Train Pass & Mean Reward & KL & Refl.\ Fix \\
\midrule
Baseline & 0.7963 & 0.8537 & 0.6340 & 0.6800 & -- & -- & -- & 0.1176 \\
A: Exec-only GRPO & \textbf{0.8049} & \textbf{0.8537} & 0.6360 & 0.6900 & 0.6385 & 0.6713 & \textbf{0.0011} & -- \\
C-light: Naive Co-Evolve & 0.7732 & 0.8232 & \textbf{0.6580} & 0.6800 & \textbf{0.6577} & \textbf{0.6930} & 0.0941 & \textbf{0.1389} \\
C-reg: Regularized Co-Evolve & 0.8000 & \textbf{0.8537} & 0.6300 & \textbf{0.7200} & 0.6288 & 0.6578 & 0.0136 & 0.0968 \\
\bottomrule
\end{tabular}
\end{table*}

Three patterns emerge. Execution-only GRPO scores highest on HumanEval pass@1, narrowly ahead of C-reg. Naive co-evolution (C-light) improves MBPP held-out pass@1 and training
pass rate but degrades HumanEval, showing the characteristic divergence pattern
where HumanEval falls and KL grows.
The C-reg configuration, with a lower auxiliary loss scale
($\lambda_{\mathrm{aux}} = 0.03$) and a stronger KL penalty ($\beta = 0.02$),
recovers performance on HumanEval to close to baseline and achieves the best
MBPP pass@5.

These results motivate the more conservative configuration used in the main experiment of the next section, which we call C-reg2: $\lambda_{\mathrm{aux}}$ lowered from 0.03 to 0.02, the
pairwise weight from 0.1 to 0.05, and $\beta$ raised from 0.02 to 0.03. The two
changes work together: a smaller $\lambda_{\mathrm{aux}}$ weakens how strongly
auxiliary SFT pulls the shared adapter, while a larger $\beta$ restrains drift
from the frozen pretrained model.

\subsection{Main Experiment: HumanEval and MBPP}
\label{sec:main-experiment}
Of the three co-evolution settings, we carry only C-reg2 forward, since C-light
and C-reg were used to select it. The main experiment runs two conditions for
up to six iterations on all 311 MBPP training problems, longer and on more data
than the preliminary study; we call these runs A-long and C-reg2-long. \textbf{A-long} is Condition~A with
$\beta = 0.02$. \textbf{C-reg2-long} is the conservative co-evolution
configuration, with $\beta = 0.03$, $\lambda_{\mathrm{aux}} = 0.02$, and the
objective parameter weights in Table~\ref{tab:aux-objectives}. A-long completed five iterations; the
sixth-iteration update did not complete due to a transient server interruption
during the GRPO backward pass. C-reg2-long completed six iterations. We use
iteration~5 as the matched-iteration comparison point because both runs have
completed checkpoints there. Table~\ref{tab:stage2-main-summary} summarizes
the results.

\begin{table*}[t]
\centering
\caption{Main experiment summary, with columns as in
Table~\ref{tab:stage1-summary}. The table includes best-checkpoint rows,
matched iteration~5 rows, and the final C-reg2 checkpoint. Bold marks the best
value in each column.}
\label{tab:stage2-main-summary}
\footnotesize
\begin{tabular}{lcccccccc}
\toprule
Model / Checkpoint & HE@1 & HE@5 & MBPP@1 & MBPP@5 & Train Pass & Mean Reward & KL & Refl.\ Fix \\
\midrule
Baseline & 0.7963 & 0.8537 & 0.6340 & 0.6800 & -- & -- & -- & 0.1176 \\
A-long best (iter 4) & \textbf{0.8159} & \textbf{0.8720} & 0.6380 & \textbf{0.7100} & 0.6486 & 0.6844 & \textbf{0.0023} & -- \\
A-long matched (iter 5) & 0.7963 & 0.8537 & 0.6360 & 0.6900 & 0.6723 & 0.7086 & 0.0024 & -- \\
C-reg2 best HE@1 (iter 1) & 0.7988 & 0.8476 & 0.6380 & 0.7000 & 0.5933 & 0.6237 & 0.0003 & 0.0606 \\
C-reg2 matched (iter 5) & 0.7707 & 0.8232 & 0.6280 & 0.6700 & 0.6574 & 0.6857 & 0.0822 & 0.0811 \\
C-reg2 final (iter 6) & 0.7744 & 0.8232 & 0.6320 & 0.6800 & \textbf{0.6962} & \textbf{0.7274} & 0.0957 & 0.0833 \\
\bottomrule
\end{tabular}
\end{table*}

Condition~A produces the best held-out HumanEval performance: pass@1 of 0.8159
and pass@5 of 0.8720 at iteration~4, with KL of 0.0023. At the matched
iteration~5 comparison point, A-long has HumanEval pass@1 of 0.7963 and KL of
0.0024, while C-reg2 has pass@1 of 0.7707 and KL of 0.0822. C-reg2 continues to
iteration~6, where training pass rate reaches 0.6962 and KL rises to 0.0957,
while HumanEval pass@1 remains below baseline at 0.7744. The selected A-long
iteration~4 checkpoint improves HumanEval pass@1 by 1.96 percentage points (pp)
and pass@5 by 1.83~pp over the pretrained baseline, and improves MBPP held-out
pass@5 by 3.00~pp, all at very low KL drift.

To avoid a misleading best-vs-final comparison, we also compare the runs at
the matched iteration. Best checkpoint to best checkpoint, A-long exceeds
C-reg2-long by only 1.7~pp on HumanEval pass@1 (0.8159 vs 0.7988); at the
matched iteration~5, A-long is back at baseline (0.7963) while C-reg2-long is
below it (0.7707). Notably, C-reg2-long achieves its highest training pass rate (0.6962)
over the same interval that HumanEval pass@1 falls. Training pass rate alone is
therefore not sufficient evidence of self-improvement: the model gets better at
the training loop, not at the held-out distribution.

Two factors limit how much weight we place on the pass@$k$ comparisons. First, HumanEval contains 164 problems and our experiments use a single random seed, so a
2--4 percentage-point gap represents only 3--7 individual problems changing between passing and failing, which is within the range of expected variation across seeds and too
small for a significance claim. Second, the pretrained baseline already achieves
0.7963 on HumanEval pass@1, leaving limited room for measurable improvement. For
these reasons we treat the KL divergence and trained-token-length diagnostics, both of which
show large and monotonic differences between conditions, as the more informative
signals, and treat pass@$k$ gaps as supporting evidence rather than the central
finding.

\begin{figure}[ht]
\centering
\includegraphics[width=\linewidth]{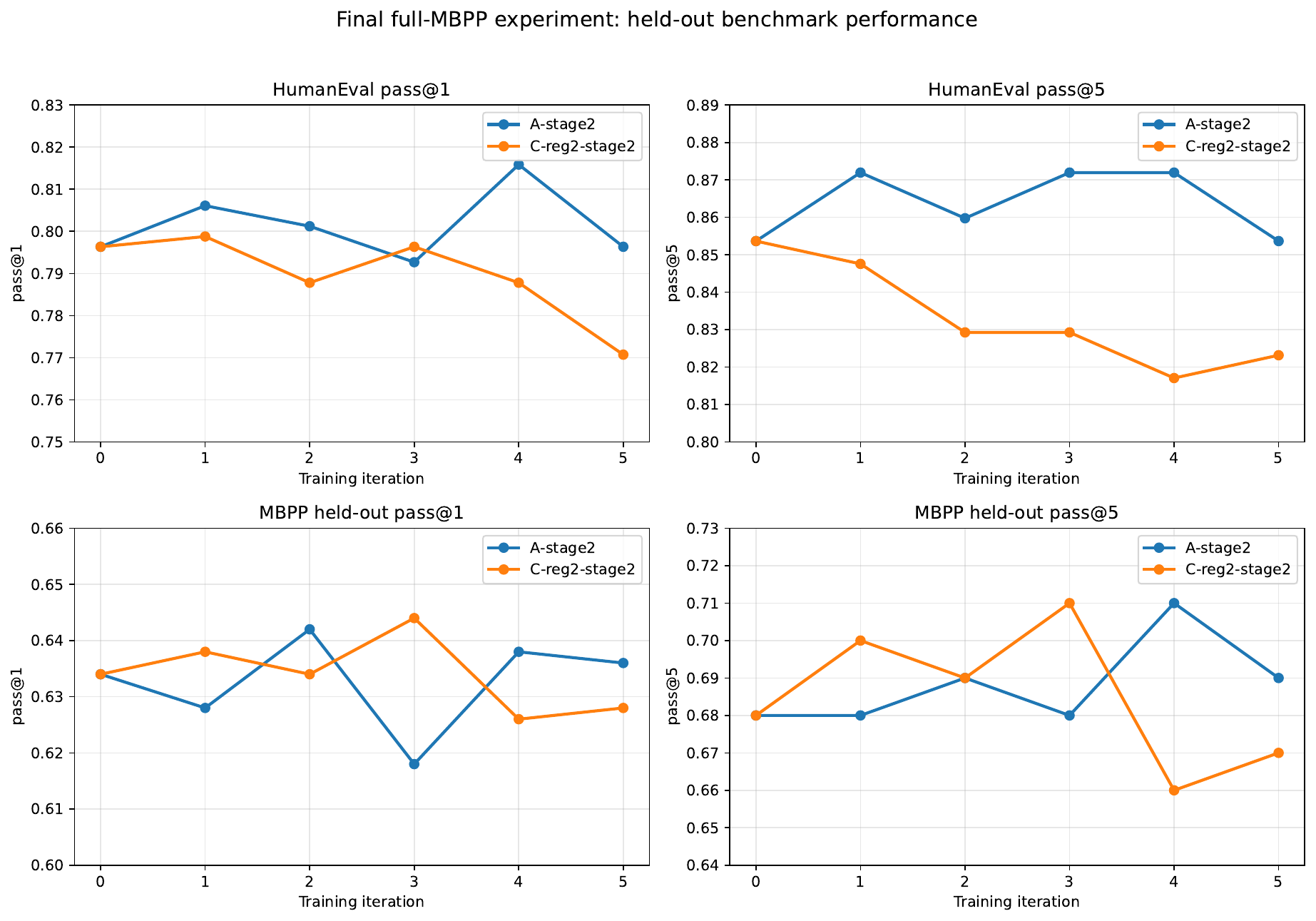}
\caption{Held-out benchmark trajectories. Condition~A achieves the strongest
HumanEval pass@1, while C-reg2 does not improve cross-benchmark generalization.
Dashed lines mark the pretrained iteration-0 baseline. Comparison is across
iterations 0--5 where both conditions have completed checkpoints. In all
figures, the runs are labeled \texttt{A-stage2} and \texttt{C-reg2-stage2} in the plots, from the internal naming of the second experiment stage; these are A-long and C-reg2-long.}
\label{fig:stage2-benchmark}
\end{figure}

Figure~\ref{fig:stage2-benchmark} shows the held-out benchmark trajectories.
Under HumanEval pass@1, A-long is the strongest condition, peaking at
iteration~4; C-reg2-long peaks at 0.7988 in iteration~1 and declines
thereafter. The MBPP held-out curves provide a complementary view: C-reg2
briefly improves MBPP performance, but the improvement is not durable. Both
conditions reach MBPP pass@5 of 0.7100 at their best checkpoints, and C-reg2
returns to baseline by iteration~6. Co-evolution can therefore temporarily
improve in-distribution MBPP metrics, but these gains coincide with policy
drift and do not transfer to cross-benchmark HumanEval.

\subsection{Mechanism: Training Improvement Versus Policy Drift}
\label{sec:mechanism}
\begin{figure}[ht]
\centering
\includegraphics[width=\linewidth]{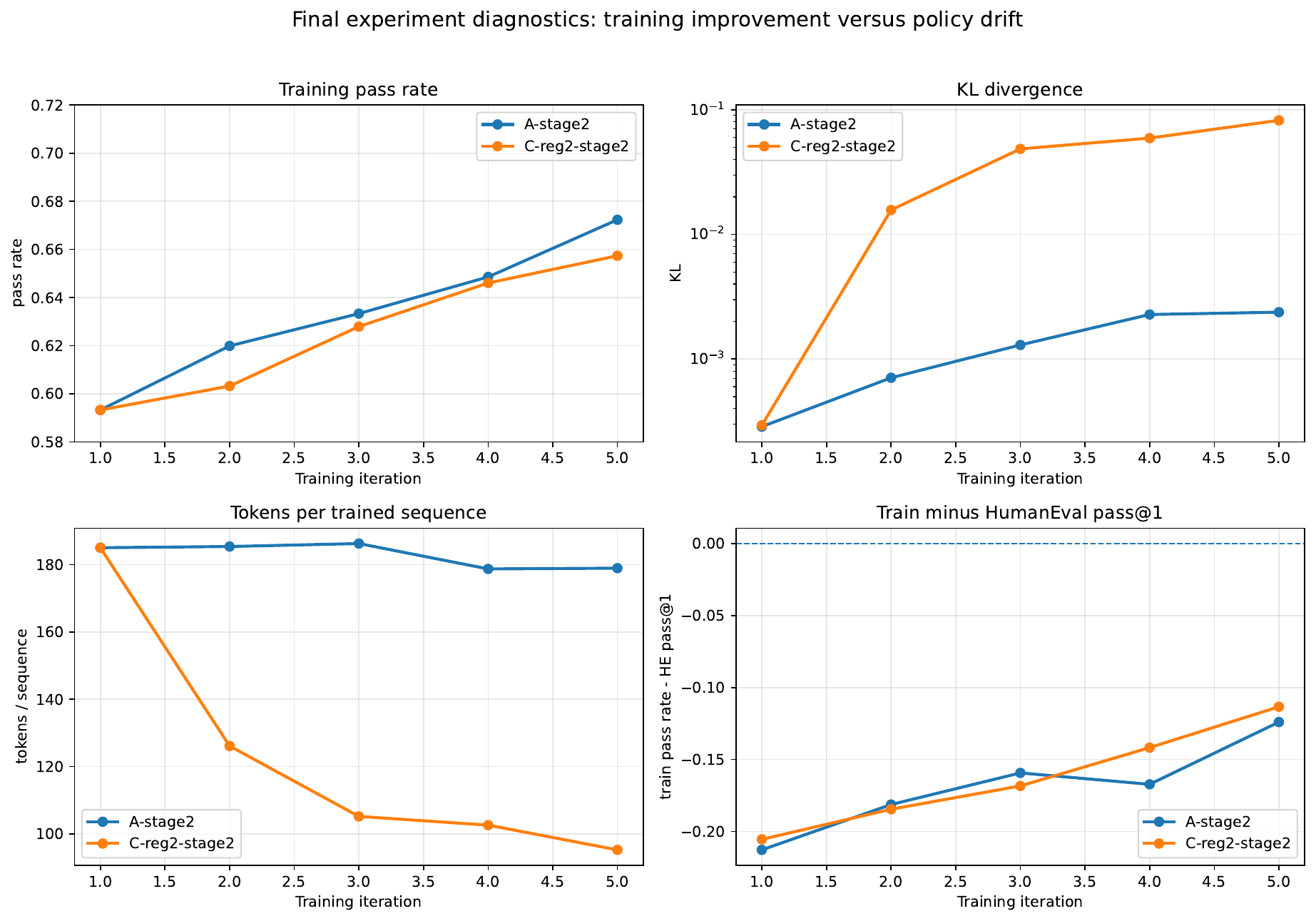}
\caption{Mechanism diagnostics. Both conditions improve training pass rate, A
slightly more than C-reg2, but C-reg2 shows much larger KL drift and a sharp
decline in tokens per trained sequence. In the train-minus-HumanEval panel,
both gaps narrow because training pass rate rises; the difference is that A's
HumanEval holds near baseline while C-reg2's falls.}
\label{fig:stage2-mechanism}
\end{figure}

Figure~\ref{fig:stage2-mechanism} shows the central divergence across
iterations 1--5; as in Figure~\ref{fig:stage2-benchmark}, the plotted series
labeled \texttt{A-stage2} and \texttt{C-reg2-stage2} are A-long and
C-reg2-long. C-reg2-long and A-long both improve training pass rate, but the gap with
held-out HumanEval evolves differently: for A-long, training pass rate rises
while HumanEval stays near baseline, whereas for C-reg2-long, training rises as
HumanEval falls. By iteration~5, A-long's KL is 0.0024 while C-reg2-long's is
0.0822, about $34\times$ larger, and the number of trained tokens per sequence
falls from 185 at iteration~1 to 95 by iteration~5 in C-reg2-long while A-long
holds near 180. By iteration~6 the trend is more extreme (KL 0.0957,
tokens 86), but the same mechanism is already visible by iteration~5.

This large drift is consistent with the HumanEval regression: even though only
0.96\% of parameters are trainable through LoRA, the auxiliary objective is
strong enough to move the model far from its pretrained behavior. The pattern
also matches recent studies of reinforcement learning for code. For example, a study of self-improving code agents~\cite{rewardhacking} found that in 73.8\% of
GPU-kernel optimization trajectories and 46.8\% of algorithmic-optimization
trajectories, the agent's reward improved without corresponding improvement on
held-out evaluation, with the gap widening as optimization continued; similar
over-optimization appears in code-repair self-play~\cite{anchoredselfplay}. We
observe the same widening pattern in our training regime.

We identify four failure signatures consistent with the observed divergence.
The first three concern the shared-adapter design itself.
(1)~\textbf{Auxiliary target heterogeneity}: pairwise outputs (single-token
A/B) and reflection outputs (full programs) differ in length and structure from
normal code completions, so when the same adapter is updated by all loss
signals it is pulled toward a compromise distribution that fits none of the
objectives well. (2)~\textbf{Length collapse}: trained tokens per sequence fall
from 185 to 86, consistent with the adapter shifting toward the shorter auxiliary targets.
(3)~\textbf{Single-adapter entanglement}: because policy and auxiliary
objectives share the same LoRA weights, the training pressure from each cannot
be balanced through architectural separation alone, even with reduced auxiliary
loss scale and increased KL regularization. The fourth reflects how we
constructed reflection targets, as discussed in the next section.
(4)~\textbf{Weak reflection supervision}: the trained 
reflection fix rate remains below the pretrained baseline at every iteration.
The baseline repairs 4 of 34 failing programs (0.1176); C-reg2-long falls to
2/33 at iteration~1 and recovers only to 3/36 (0.0833) by iteration~6. These
counts are small enough that we read the metric as directionally consistent
with the other diagnostics rather than as an independent result.

The progression across configurations (C-light $\rightarrow$ C-reg
$\rightarrow$ C-reg2) strengthens this reading. Naive C-light caused immediate
HumanEval degradation and large KL growth; C-reg reduced the damage and
improved MBPP pass@5; and the more conservative C-reg2 still drifted under more
data and longer training. All three showed drift away from the pretrained
distribution. In the preliminary study, C-light, with a stronger auxiliary loss and a weaker KL penalty than C-reg, drifted far more; C-reg2 still drifted
substantially despite the weakest auxiliary loss of the three, though it was
also trained longer and on more data. This consistency across regularization regimes suggests the divergence is
intrinsic to single-adapter co-evolution under these objectives, not an
artifact of a single poorly chosen configuration.

%-------------------------------------
\section{Discussion and Conclusion}
%-------------------------------------
SPARK-style co-evolution is applicable to code generation, but it does not outperform execution-only GRPO on cross-benchmark HumanEval. At the matched
iteration~5, C-reg2 trails A-long on HumanEval pass@1 (0.7707 versus 0.7963)
while its KL divergence is about $34\times$ larger. The co-evolution
branch receives strictly more information than Condition~A, including paired
comparisons and repair targets, yet because all
objectives update the same shared adapter, auxiliary learning interferes with
the generation policy. Training pass rate should therefore be treated as a
diagnostic, not a final quality metric. These pass@$k$ magnitudes are
single-seed trends rather than significance claims; the KL and length-collapse
diagnostics, which are large and monotonic, are the more robust signal.

Single-adapter co-evolution may be too architecturally entangled for the code
setting. Code generation breaks the structural symmetry SPARK relies on in
reasoning tasks: in math and visual reasoning, rollout, judgment, and
reflection outputs all resemble a reasoning trace followed by an answer,
whereas in code, policy rollouts are full programs, pairwise targets are single
tokens, and reflection targets are full code conditioned on a different prompt.
Training one shared adapter on this heterogeneous mixture is a harder
distribution-balancing problem, and the observed length collapse is consistent
with that mismatch. 

A caveat applies to how strongly this speaks to co-evolution in general. Our
auxiliary mixture is reflection-dominated, and reflection uses canonical or
correct-rollout targets rather than verified self-fix trajectories. The length
collapse and below-baseline reflection fix rate are therefore partially
entangled: we cannot fully separate heterogeneous-target drift from weak
reflection construction. The failure we document is for a reflection-heavy,
canonical-target mixture co-updating a single adapter, not for all co-evolution
weightings. Our result aligns with SCoRe's distribution-shift and
behavior-collapse predictions~\cite{score}, and with the pattern reported
concurrently by Choi et al.~\cite{anchoredselfplay} in code-repair self-play,
where generator-fixer training improves repair on self-generated bugs while
degrading repair on human-authored bugs.

Several prior approaches give code-generation models a reasoning or explanation
channel without SPARK-style co-evolution. Chain-of-thought, Self-Debug, and
Reflexion use prompting, execution feedback, or verbal reflection at inference
time or in separate training formats~\cite{cot,selfdebug,reflexion}. Concurrent
code self-improvement methods factor the policy and auxiliary roles
differently. ACE~\cite{ace} trains the model both to solve problems and to generate
adversarial unit tests that expose flaws in its own solutions, while GASP
(Guided Asymmetric Self-Play) uses a teacher-student setup in which one role
poses problems and the other solves them~\cite{gasp}. Both report improved code-generation performance from self-improvement, which is consistent with our interpretation that the
limitation lies not in self-improvement on code in general, but in the design
choices studied here, particularly co-updating a single shared adapter with
heterogeneous objectives.

Within co-evolution, separating the policy adapter from a dedicated auxiliary
adapter on a shared base model is the most promising direction we see, and we
are not aware of prior work testing it for code generation. Methods that avoid
co-evolution entirely, such as preference optimization from execution-labeled
pairs, are a separate direction. Better reflection
may also require verified repair trajectories and reference-anchored fixer
training~\cite{anchoredselfplay}, together with held-out monitoring to avoid
proxy-reward over-optimization~\cite{rewardhacking}. More broadly, even when
feedback comes from executable unit tests rather than a learned reward model,
co-evolving generation and evaluation within one adapter can produce training
gains that do not transfer. Stronger supervision alone may therefore not be sufficient; architectural separation between generator and evaluator may also be needed.

\section*{Limitations}
This study evaluates a single model at a single scale using a single random
seed over six iterations, so the pass@$k$ differences we report are trends
rather than significant effects. HumanEval is also a widely used benchmark, so
we cannot rule out that some of its problems appeared in the pretraining data
of the model we fine-tune.

Because the co-evolution condition adds an auxiliary update phase, it takes
more optimizer steps than the baseline. We did not run a step-matched
comparison, in which the baseline receives the same number of updates, so we
cannot separate the effect of additional training from that of the auxiliary
objectives. We also do not compare against ACE~\cite{ace} or
GASP~\cite{gasp}, which factor the policy and auxiliary roles differently.
Finally, our auxiliary mixture is dominated by reflection, whose weight is
twenty times that of pairwise comparison. The results therefore do not speak to
co-evolution under other weightings.

\bibliographystyle{IEEEtran}
\bibliography{references}

\end{document}